%% file: main.tex
\documentclass[runningheads,orivec]{llncs}
\usepackage[T1]{fontenc}
\usepackage{graphicx}
\usepackage{amsmath}
\usepackage{tikz}
\usepackage{pgfplots}
\pgfplotsset{compat=1.18}
\usepackage{subcaption}
\usepackage{cleveref}

\newcommand{\Chart}[3]{#1~
    \begin{tikzpicture}
        \fill[color=black!10] (0, 0) rectangle (7ex, 1.5ex);
        \fill[color=black!30] (0 , 0) rectangle ({((#1-#2)/(#3-#2))*7ex}, 1.5ex);
    \end{tikzpicture}}
    
\newcommand{\ChartH}[3]{\textbf{#1}
    \begin{tikzpicture}
        \fill[color=black!10] (0, 0) rectangle (7ex, 1.5ex);
        \fill[color=green!30!black!50] (0 , 0) rectangle ({((#1-#2)/(#3-#2))*7ex}, 1.5ex);
    \end{tikzpicture}}

\begin{document}
\title{Distributional Loss for Robust Classification}
%
%
\author{Kathleen Anderson\inst{1} \and Thomas Martinetz \inst{1} }
%
\authorrunning{K. Anderson and T. Martinetz}
%
\institute{Institute for Neuro- and Bioinformatics, University of Lübeck, Germany\\
\email{\{k.anderson,martinetz\}@uni-luebeck.de}}
\maketitle              
\begin{abstract}
This paper proposes a novel loss concept for supervised classification tasks. Rather than enforcing a direct mapping from each input sample to a single assigned label, we define an optimization objective over all classifier outputs as a bimodal Gaussian distribution. This softer target formulation implicitly captures class ambiguity, mitigates overfitting, and encourages the learning of more robust decision boundaries, all without requiring additional label information. Experimental results demonstrate consistent improvements in robustness, with particularly pronounced gains in low-data regimes, while requiring only minimal modifications to standard training pipelines.

\end{abstract}
\section{Introduction}
The domain of robust classification strategies is vast and spans a wide range of different approaches, including dataset regularization or augmentation, network architectures, and pretraining. An aspect whose relevance has diminished in recent years is the choice of classification training loss, with cross entropy often serving as the gold standard for any classification task. 

In this work, we consider a supervised classification setting in which inputs 
$x \sim P(x)$ are associated with class labels 
$c_k \sim P(c_k \mid x)$, where $k = 1, \dots, K$ denotes the $K$ possible classes to which an input can be assigned. In the idealized case, if $P(c_k \mid x) = 1$, the input $x$ always belongs to class $c_k$ with certainty. Conversely, if $P(c_k \mid x) = 0$, the input $x$ will never belong to class $c_k$. These probabilities of 0 and 1 represent extreme cases. In practice, due to noise, class overlap, or shared features among different classes, posterior probabilities are rarely exactly $0$ or $1$. Instead, they typically take values strictly between these extremes.   

Most loss functions, such as the cross-entropy loss, are nevertheless defined with respect to these extremes. In a training dataset $\{(x^{(n)},c_k^{(n)}), n=1,...,N\}$, to each input $x^{(n)}$ is assigned a hard class label $c_k^{(n)} \in \{c_1, \dots, c_K\}$. This corresponds to a target distribution for which $P(c_k^{(n)} \mid x^{(n)})=1$ for the correct class and $P(c_l^{(n)} \mid x^{(n)})=0$ for all $l \neq k$. As a result, the classifier is effectively trained to reproduce these degenerate (one-hot) target distributions.

Assigning each sample a single definitive label may fail to capture the intrinsic complexity of the underlying task. In many practical scenarios, samples are not perfectly representative of a single class, but instead exhibit characteristics shared across multiple classes. Using softer, more informative labels, ideally reflecting the true class probabilities, should improve classification robustness \cite{Related_2014_ClassificationModelsWith}, but such high-quality probabilistic annotations are rarely available in practice. Enforcing a rigid, hard label assignment instead may encourage overfitting by compelling the model to learn overly confident and potentially misleading decision boundaries.

This work utilizes the assumption that for most natural datasets the distribution of the posterior probability values $p_k(x)=P(c_k \mid x)$ for inputs $x \sim P(x)$, is more likely to follow a bimodal distribution as illustrated in \cref{tikz:target_dist}, with modes concentrated near $0$ and $1$.

We propose a \textit{distributional loss} that explicitly leverages this bimodal structure. The training labels now indicate only to which mode of the distribution each sample should be assigned. The proposed training objective encourages the classifier to distribute predictions in accordance with this bimodal behavior. To achieve this, we employ two complementary differentiable estimators of the Kullback--Leibler divergence. The first is based on a \textit{probabilistic encoder}, as commonly used in variational autoencoders, while the second relies on a non-parametric density estimator derived from pairwise distances between samples. These two loss components complement each other, with one creating a robust base and the other fitting more closely to the true distribution.

Experimental results demonstrate improved robustness, particularly in settings with limited training data. The proposed objective is especially advantageous in regimes where the risk of overfitting is high. The required modifications to the training procedure are minimal and can be readily integrated into virtually any classifier architecture or training pipeline.

\section{Related Work}
\label{sec:related_work}

Cross entropy in a combination with a softmax continues to serve as the standard loss for neural network classification in contemporary research \cite{Related_2025_ComprehensiveSurveyOf}. Contradicting its omniscience, previous research has identified various shortcomings of this loss function, and the idea of augmenting or replacing cross entropy has been approached from many angles.

One of the discovered disadvantages is a lacking resilience to noise or outliers. It has been shown that contrary to cross entropy, the absolute distance between network output logits and target label is robust to label noise \cite{LossFunctions_2017_LossFunctionsUnder}, which motivated the proposal of new loss functions that enable smooth transitions between cross entropy and absolute distance \cite{LossFunctions_2018_CrossEntropyLoss}, based on a scalar hyperparameter. 

Another caveat of cross entropy training is that it has been found to motivate overconfidence - the output of a trained classifier almost always claims a very high probability (i.e. a high confidence) for one class and a very low probability for all others, even when the selected target class is incorrect. An added confidence penalty can improve the classification accuracy \cite{LossFunctions_2017_NeuralNetworksBy,LossFunctions_2020_TrainingSpeedAccuracy}.

The application of softmax implicitly enforces margins between class logits, but depending on the dataset, the resulting margin was shown to be inconsistent or insufficient. Explicitly increasing the margin between class logits can yield more robust classifiers \cite{LossFunctions_2016_SoftmaxLossFor,LossFunctions_2016_LargeMarginDeep,LossFunctions_2016_TheDepthOf,LossFunctions_2019_ImbalancedDatasetsWith,LossFunctions_2018_MarginDeepNetworks,LossFunctions_2018_OrthogonalLow-rankEmbedding,LossFunctions_2018_FeatureDistributionFor}. On a more theoretical basis, another line of research reported findings about a reweighted Taylor-decomposition of standard cross entropy loss \cite{LossFunctions_2021_APolynomialExpansion,LossFunctions_2021_LossFunctionsThrough}.

Previous research has also, like this paper, pondered on the idea of replacing cross entropy entirely. Common alternatives include variations of cosine similarity \cite{LossFunctions_2020_LearningOnSmall,LossFunctions_2020_VisualInductivePriors,LossFunctions_2021_SimilarityForRegularizing} and correntropy \cite{LossFunctions_2014_C-lossFunctionFor,LossFunctions_2017_DCNNByCombining,LossFunctions_2024_Correntropy-basedRobustDistance}. Correntropy in its basic form denotes a simple kernelized distance between classifier output and target vector: $L_{corr}(x, y) = \kappa(x - y)$, with $\kappa$ usually defined to be a Gaussian. Empirical exploration yielded other entirely new (and in some cases even counter-intuitive) loss functions \cite{LossFunctions_2019_GeneralAndAdaptive,LossFunctions_2017_EntropyCostFunction}, which also outperform their cross entropy baseline. 

Naturally, specialized loss functions have also been employed for a multitude of other purposes, such as the search for robust representations \cite{LossFunctions_2018_GeneralizationViaScalable,LossFunctions_2019_AndImprovingRepresentations,LossFunctions_2020_ContrastiveLearning.md,LossFunctions_2021_NegativeSamplingFor} or the regularization of intermediate feature layers \cite{LossFunctions_2024_AndEnhancingThe}.

\section{Training for a Distribution}
\label{sec:prob_training}
Following the general considerations outlined in the introduction, we now turn to a concrete setting. We consider a classifier implemented as a neural network with $K$ output units, one for each class. 

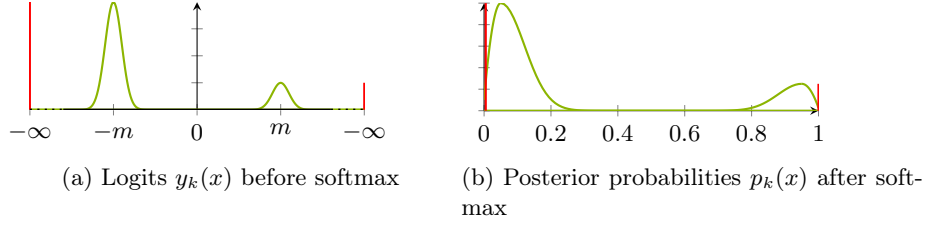
\begin{figure}
    \input{Figures/target_dist}
    \caption{Target for of a single output unit (i.e. corresponding to a single class), for a standard classification task, defined by softmax cross entropy (red) and our new formula (green).}
    \label{tikz:target_dist}
\end{figure}

When using a softmax layer, the outputs of the classifier are interpreted as posterior probabilities $p_k(x) = P(c_k \mid x)$ for a given input $x$. Let $y_k(x)$ denote the corresponding pre-softmax activations (logits). In the typical classification scenario, all target probabilities are given as either $p_k(x)=1$ or $p_k(x)=0$ (marked red in \cref{tikz:target_dist}). While the probabilities $p_k(x)$ are constrained to the interval $(0,1)$, the logits $y_k(x)$ take values in $(-\infty, \infty)$, with $y_k(x) \to -\infty$ for $p_k(x) \to 0$ and $y_k(x) \to \infty$ for $p_k(x) \to 1$. The scalar values $y_k(x)$ will henceforth be abbreviated as $y$ when the indices are clear.

Our new distributional target is based on the assumption that for most datasets, a softer curve is closer to the ground truth (drawn in green in \cref{tikz:target_dist}). Instead of aiming for $\infty$ and $-\infty$, the logit values $y$ are aimed at a bimodal Gaussian distribution with means $-m$ and $m$. 

Contrary to cross entropy, our new loss does not require its inputs to be probabilities and therefore does not include a softmax layer, but is computed directly on the logits $y$. The fixed target distribution nonetheless implicitly constrains the range of output values. A notable effect of omitting softmax is that $y_k(x)$ for one $x$ do not directly influence each other: a high $y_k(x)$ for one class $c_k$ does not necessarily lower the value of the other $y_l(x), l \neq k$. 

To minimize the divergence between the newly defined target distribution and the output of the classifier, we use the Kullback-Leibler (KL) divergence and two strategies of computing it: a parametric solution that is robust, but very strict, and a non-parametric estimator which is more sensitive to noise, but adapts more flexibly to a distribution. The two losses are combined with a balancing factor $\lambda$ as
\begin{equation}
    L(y) = \lambda L_{\text{param}}(y) + L_{\text{pointwise}}(y, Y)
\end{equation}
where the pointwise loss $L_{\text{pointwise}}$ is computed for the logit value $y$ in relation to the set $Y$ of all $y$.

\subsection{Parametric Probabilistic Encoders}
\label{sec:prob_enc}

A probabilistic encoder is usually used as a component of a variational autoencoder (VAE). Instead of directly computing the logit $y(x)$ for an input $x$, the output of a probabilistic encoder $f$ consists of the parameters of a Normal distribution, $\mu_x$ and $\sigma_x$. Analogously to a VAE, we define the target to be another Normal distribution $\mathcal{N}(\mu_{\text{target}}, \sigma_{\text{target}})$, and compute the loss via the closed form equation for the KL divergence
\begin{equation}
L_{\text{param}}(\mu_x, \sigma_x) = KL(\mathcal{N}(\mu_x, \sigma_x) \ | \  \mathcal{N}(\mu_{\text{target}}, \sigma_{\text{target}})).
\end{equation}
The standard deviation $\sigma_{\text{target}}$ is shared between all inputs, while $\mu_{\text{target}}$ is defined as $+m$ or $-m$ depending on the class assigned to input $x$, resulting in the bimodal Gaussian distribution depicted in \cref{tikz:target_dist}.

\subsection{Generating Samples from a Probabilistic Encoder}
A probabilistic encoder does not generate logits, but distributions over logits, each defined by their parameters $\mu_x$ and $\sigma_x$. The pointwise divergence estimation introduced in the next section, however, is based on a fixed set of points. To utilize the logits $y$ itself, one can implement the reparameterization trick: draw a random $n \in \mathcal{N}(0, 1)$ and compute $y(x) = \sigma_x n + \mu_x$. Gradient descent can be done as if $y$ was directly generated by $f$.

The reparameterization trick comes with another advantage: the logits can be augmented by drawing several $y(x)$ per input $x$. Since the pointwise divergence estimation in the next section requires many data points to work well, this natural dataset expansion can be especially beneficial.

\subsection{Non-Parametric Pointwise Divergence}
\label{sec:nonpar_KL}
The outcome of the probabilistic encoder, after reparameterization, is a set of logit values $y$ drawn from many Normal distributions, with one individual distribution $\mathcal{N}(\mu_x, \sigma_x)$ corresponding to every datapoint $x$. The KL divergence loss described in the previous section aims to make all of those $\mathcal{N}(\mu_x, \sigma_x)$ match one of two targets, $\mathcal{N}(+m, \sigma_{\text{target}})$ or $\mathcal{N}(-m, \sigma_{\text{target}})$. However, the combined set of all $y$ can reflect a target distribution even if individual points originate from distinct $\mathcal{N}(\mu_x, \sigma_x)$. In fact, it can be beneficial to allow deviations for some $x$, e.g. to make the classifier respond with a low confidence to a noisy sample.

The second divergence estimator balances the strict goal of the probabilistic encoder with a non-parametric density estimation. Instead of a purely sample-wise evaluation, the estimator approximates the distribution over an entire batch of logits. As before, the scalar logit values $y$ can be split into two categories: 
\begin{description}
    \item[positive] when the class $c$ assigned to $x$ corresponds to the output unit, aiming for $\mathcal{N}(+m, \sigma)$. The standard softmax cross entropy formulation would have trained for $y \to \infty$ for all $y$ in this group (see \cref{tikz:target_dist}).
    \item[negative] $y$ are those where the logit dimension does not match the target class, which aim for an overall distribution of $\mathcal{N}(-m, \sigma)$. 
\end{description}

This assignment yields a total of $2 K$ groups for $K$ classes: one positive and one negative group for each of the $K$ classifier output units. Let $Y_l \subset Y$ be a set of $y$ belonging to the same group and aimed at the same target distribution. For each $Y_l$, we sample a matching set $T_l$ of points from the assigned target distribution\footnote{While the explicit estimation of density values is also possible, we found this more indirect estimation of comparing two sets of points to yield better results}, $|Y_l| = |T_l|$. 

As a minimization objective proportional to the KL--divergence between the distributions $P(Y)$ and $P(T)$, we insert one of the subsequently described divergence measures $d$ into 
\begin{equation}
    L_{\text{pointwise}}(y, Y) = d(y, Y_l),
\end{equation}
with $Y_l \subset Y$ referring to the set of all scalar logit values that correspond to the same group as $y$. 

\subsubsection{Nearest-Neighbor Estimation}
The nearest neighbor density estimation method is based on a simple intuition: for each $y \in Y_l$, estimate the distance between $y$ and its neighbors. A close proximity to many neighbors indicates a high density area. If the distributions of two sets $Y_l$ and $T_l$ align, the average neighbor distances for a point at the location of $y$ will be the same when estimated in either $Y_l$ or $T_l$. The strategy was first used in the context of entropy estimation \cite{kozachenko1987knnDensity} and has only recently been applied to gradient based learning \cite{anderson2026nearest}. 

For a value $y$, let $\{y^{(1)}, ..., y^{(m)}\}$ and $\{t^{(1)}, ..., t^{(m)}\}$ be the $m$ closest neighbors to $y$ in $Y_l$ and $T_l$ respectively, sorted by their distance $\Delta(y, y^{(i)})$. The KL divergence between $P(Y_l)$ and $P(T_l)$ can be decreased by minimizing
\begin{equation}
    d_\text{nn}(y, Y_l) = \frac{1}{m} \sum_{i = 1}^{m} (\Delta(y, y^{(i)}) - \Delta(y, t^{(i)}))^2.
\end{equation}

Interestingly, $d_\text{nn}$ is not only proportional to the true KL divergence, but can be reshaped into an accurate approximation of it, with the approximation error decreasing with the smoothness of the true distribution. Refer to \cite{anderson2026nearest} for a more in-depth derivation and analysis.

The hyperparameter $m$ heavily influences the accuracy of the $d_\text{nn}$ estimator: considering only very close neighbors can make the estimate sensitive to noise and too uneven, choosing a very large $m$ may oversmooth the estimate and obscure divergence. To ease the influence of $m$, we chose $\Delta(y, y^{(i)}) = e^{-(y - y^{(i)})^2}$. The Gaussian kernel weights close neighbors higher, but still factors distant neighbors into the difference.

\subsubsection{Kernel Density Estimation}
When computing the distributional divergence on one-dimensional data, the popular kernel density estimation (KDE) is exceedingly similar to nearest-neighbor estimation, the main difference being that the density is estimated as the kernelized sum over all points first, and compared second:
\begin{equation}
    d_\text{kde}(y) = \left(\sum_{i = 1}^{|Y_l|}\Delta_h(y, y^{(i)}) - \sum_{i = 1}^{|Y_l|}\Delta_h(y, t^{(i)})\right)^2.
\end{equation}
This rearrangement makes the ordering of $\{y^{(1)}, ..., y^{(m)}\}$ and $\{t^{(1)}, ..., t^{(m)}\}$ irrelevant. It is also less sensitive: even when point distances diverge significantly for singular neighboring points, the sum over all neighbors may still yield two similar density estimates. 

The traditional implementation considers all points as "neighbors" ($m=|Y_l|$), eliminating the hyperparameter $m$ in favor of another hyperparameter: $h$, the bandwidth of the kernel $\Delta_h(a) = e^{-\frac{a}{h}^2}$. As the name suggests, $h$ determines the width of the kernel, and effectively decides how distant neighbors factor into the result. 

The choice of $h$ is critical for the accuracy of KDE. Unlike standard applications, manually tuning $h$ is ineffective for our purpose, as the dataset is subject to constant change over the course of the training. Our simple solution is to combine KDE and nearest neighbor strategies and scale $h$ by the average distance between a point and its closest neighbors.

\section{Distributional Loss Compared to Cross Entropy}

As demonstrated in numerous prior studies (see \cref{sec:related_work}), the standard cross-entropy loss exhibits several limitations, particularly in settings with small or low-quality datasets. The proposed distribution-based loss addresses many of these shortcomings, thereby increasing the likelihood that the learned model better approximates the underlying true data-generating distribution.

One beneficial property it shares with many related papers is the ability to explicitly define and enforce margins between classes. Since the target mean for the negative and the positive group can be chosen freely, the margin between the two is an integral part of the loss formulation.

Countering the presumed tendency towards overconfidence \cite{LossFunctions_2017_NeuralNetworksBy,LossFunctions_2020_TrainingSpeedAccuracy}, our target distributions do not only allow, but even enforce logit vectors with a lower confidence. A small percentage of classification results is expected to be uncertain, reflecting the true distribution of most datasets. Noisy samples are not forcibly treated as perfect representatives of their class.

A less obvious advantage is the independence between the dimensions of a logit vector. The classification problem assigns exactly one class to each sample, but a single input might resemble multiple classes (e.g. an image of a distant four legged animal within an animal classification dataset) or none at all (e.g. an unclear image lacking any identifiable animal). The independent logit evaluation allows such cases: a sample can be given a high score for more than one class. Nonetheless, the classifier is trained for a globally normalized solution: the output distribution over the full dataset should remain constant.

Concluding, it can be said that training with the distributional loss introduces a more intricate bias for the labeling of the dataset. Instead of mapping each sample to one class index, the loss defines properties over the entire set of outputs. Our experiments have shown that this target is generally more difficult to train for, demanding more training time and more careful tuning of hyperparameters to locate a satisfactory loss minimum. However, the assumption of a smoother label distribution is usually much closer to the true classification problem. Therefore, once a good training accuracy is achieved, the result is more likely to generalize well.

\section{Experiments}
We compare our new method against several other loss functions. Softmax cross entropy (CE) serves as a baseline, and results are also presented for \textit{PolyLoss}  \cite{LossFunctions_2021_APolynomialExpansion}, a popular variation of cross entropy that reweights the first term of its Taylor Expansion, and \textit{T-vMF} Loss \cite{LossFunctions_2021_SimilarityForRegularizing}, which modifies the final network layer to regularize penultimate features using a generalized cosine similarity.

%
Unless stated otherwise, all experiments have been performed using the same standard ResNet18 \cite{resnet} architecture with initial weights obtained by pretraining on ImageNet. For a clean comparison, we otherwise abstain from using advanced pretraining, augmentation or normalization techniques. 

To offer a perspective on different types of image data, experiments are performed for four datasets: Cifar10 \cite{Cifar10} is one of the most commonly used general image classification datasets, featuring 60000 32x32 images split into 10 balanced classes. Caltech256 \cite{Caltech256} is a more challenging dataset comprising 30607 images from 256 diverse categories. The dataset is not balanced, each class is represented by a minimum of 80 and a maximum of 827 images. Images have been rescaled to 256x256. We also evaluate BloodMNIST: a set of 17091 microscopic images of individual blood cells divided into 8 categories, and PathMNIST: 107180 stained histological images from 9 different types of tissues. BloodMNIST images originate from the MedMNIST dataset \cite{medmnistv1} and have been scaled to 28x28. PathMNIST was published in an expanded version with larger images \cite{medmnistv2} and our experiments are performed on image size 128x128. These two medical datasets deviate more pronounced from the pretraining distribution.

For the non-parametric component of the distributional loss (\cref{sec:nonpar_KL}), results are shown for the divergence estimate that proved most effective for the respective dataset: KDE for BloodMNIST and Caltech256, and nearest-neighbor densities for Cifar10 and PathMNIST. 

For the full implementation including all training details and hyperparameters, we refer to the public github repository\footnote{https://github.com/ka-anderson/distributional-classification-loss}.

\subsection{Classification Accuracy}
\label{sec:exp_acc_full}
\begin{table}[t]
\centering
\caption{Average classification test accuracies in \% for different dataset sizes.}
\label{tab:acc_small}
\begin{tabular}{ll|llll}\hline
            &                   & \multicolumn{4}{l}{Samples per Class} \\
           & & 2                         & 20    & 200  & full dataset \\ \hline
Cifar10    & CE               & \Chart{23.50}{15}{35}     & \Chart{47.35}{40}{60} & \Chart{67.13}{60}{80} & \Chart{83.28}{80}{90} \\
           & Distributional    & \ChartH{27.94}{15}{35}     & \ChartH{55.23}{40}{60} & \ChartH{71.20}{60}{80}  & \ChartH{84.42}{80}{90}\\
           & PolyLoss          & \Chart{25.12}{15}{35}     & \Chart{48.17}{40}{60} & \Chart{67.43}{60}{80} & \Chart{83.49}{80}{90} \\
           & T-vMF             & \ChartH{27.90}{15}{35}     & \Chart{52.73}{40}{60} & \Chart{68.92}{60}{80} & \Chart{83.40}{80}{90} \\ \hline
Caltech256 & CE               & \Chart{37.98}{30}{50}     & \ChartH{75.01}{65}{85}&  & \ChartH{83.38}{75}{85}     \\
           & Distributional    & \Chart{39.11}{30}{50}     & \Chart{73.27}{65}{85}&  &\Chart{80.91}{75}{85}    \\
           & PolyLoss          & \Chart{38.32}{30}{50}     & \Chart{74.08}{65}{85}&  & \Chart{83.34}{75}{85}     \\
           & T-vMF             & \ChartH{40.89}{30}{50}     & \Chart{69.37}{65}{85}& & \Chart{75.42}{75}{85}      \\ \hline
BloodMNIST & CE               & \Chart{44.63}{43}{63}     & \Chart{73.35}{70}{90} & \Chart{90.06}{80}{100}& \Chart{95.40}{90}{100} \\
           & Distributional    & \ChartH{61.82}{43}{63}     & \ChartH{83.14}{70}{90} & \ChartH{92.32}{80}{100}& \ChartH{96.71}{90}{100}      \\
           & PolyLoss          & \Chart{51.63}{43}{63}     & \Chart{78.65}{70}{90} & \ChartH{92.14}{80}{100} & \Chart{95.29}{90}{100}     \\
           & T-vMF             & \Chart{55.45}{43}{63}     & \Chart{79.90}{70}{90} & \ChartH{92.01}{80}{100} & \Chart{96.05}{90}{100}    \\ \hline
PathMNIST  & CE               & \Chart{56.29}{45}{65}     & \Chart{83.50}{72}{92} & \Chart{94.72}{80}{100} & \ChartH{99.63}{90}{100}\\
           & Distributional    & \Chart{59.76}{45}{65}     & \ChartH{90.04}{72}{92} & \ChartH{96.06}{80}{100}& \ChartH{99.61}{90}{100}      \\
           & PolyLoss          & \ChartH{61.19}{45}{65}     & \Chart{75.35}{72}{92} & \Chart{95.04}{80}{100}& \Chart{99.53}{90}{100}      \\
           & T-vMF             & \Chart{56.94}{45}{65}     & \Chart{88.28}{72}{92} & \Chart{95.44}{80}{100} & \Chart{99.43}{90}{100}    \\ \hline
\end{tabular}
\end{table}

As it is well known, changing the training objective function can notably affect the test accuracies of a model. Visible in \cref{tab:acc_small}, the distributional loss generally increases the test accuracy, achieving the most notable improvements on small datasets. The strongest increase can be observed on BloodMNIST, where the distributional loss outperforms cross entropy by up to 17.2 percentage points. The only dataset not improved by our method is Caltech256, where cross entropy tends to outperform both our newly introduced loss and related work. One can hypothesize that the large 256 dimensional logit vectors fall into a different task category than those of a 10- or 8- class problem. Adapting the distributional loss to this slightly different setting remains future work.

\subsection{Overfitting to Random Noise}
Theoretically, a neural network with a fixed architecture will always have the same capacity: it is able to represent a specific collection of functions, independently of the training method. In practice, however, the loss function defines the topology of the optimization landscape. A subset of solutions might be theoretically realizable by a network, but not discovered by the training procedure.

As a simple evaluation of the effect of a loss function on the practical complexity of a model, we train a classifier on an artificially generated dataset. Input samples have been generated as Gaussian noise, categorical target labels are drawn from a uniform distribution. There is no statistical relationship between inputs and labels, hence the test accuracy is always equivalent to random guessing. Any training accuracy exceeding random chance indicates that the model has memorized a direct sample–label mapping.

\begin{figure}[t]
\includegraphics[width=\textwidth]{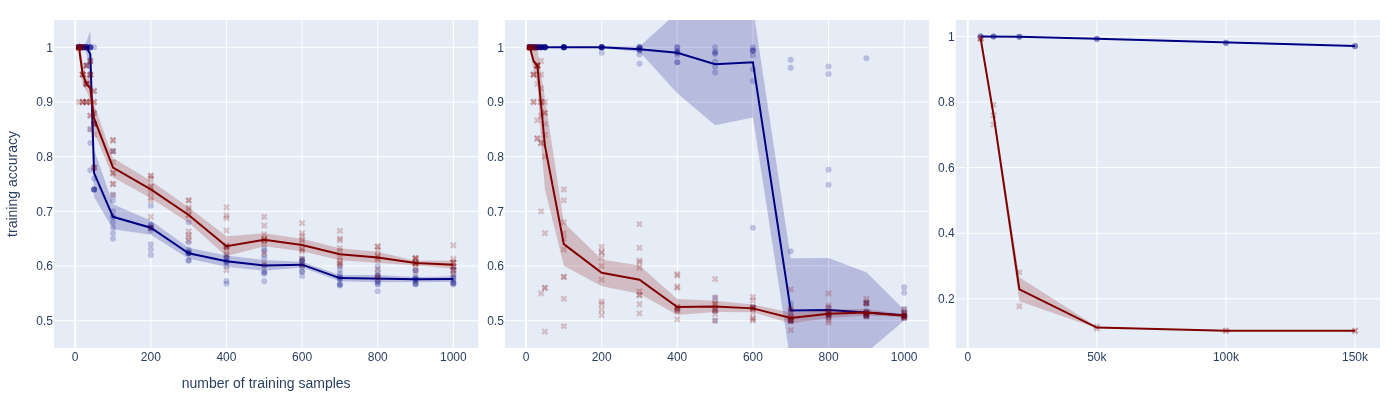}
\caption{Training accuracy of a linear classifier, and MLP and a resnet on random noise. Blue circles show models trained with cross entropy, red crosses models trained with the distributional loss.} 
\label{plot:random}
\end{figure}

The plots in \cref{plot:random} show the \textbf{training} accuracy of a linear classifier, a six layer MLP, and a small Resnet. MLP and linear classifier have been trained to learn binary labels on 20 dimensional random vectors, the Resnets are given samples with shape 30x30, attempting to arrange them into 10 classes.

The distributional loss clearly limits the memorization ability, whereas cross entropy can memorize an almost perfect training score for all dataset sizes. While it does not guarantee an increased test accuracy, a lower likelihood of memorizing random labels makes it more likely that a classifier is forced to learn meaningful structures.

\subsection{Ablation Study}
To evaluate the relevance of all loss components, we repeat our Cifar10 experiments using (i) only the parametric closed-form loss term, (ii) only the nonparametric pointwise loss term and (iii) the pointwise loss term, with a target standard deviation $\sigma=0$. This final experiment is limited to the pointwise term, since the parametric KL divergence is not defined for $\sigma=0$ and highly unstable for very small $\sigma$.

The resulting accuracies in \cref{tab:acc_ablation} demonstrate a clear benefit of using the parametric and the non-parametric loss term concurrently. Interestingly, enforcing a non-zero variance is especially important for smaller datasets: when trained on the full Cifar10 training set, the pointwise loss achieves the same accuracies, independent of target variance. However, when using 2 samples per class, the model trained for zero variance barely outperforms the cross entropy baseline. One can assume that large datasets implicitly enforce a smoother target distribution - if a certain subject $A$ can be classified as either class $i$ or class $j$, a very large dataset will contain multiple views of $A$, being labeled as either $i$ or $j$, so that minimal training error is achieved by predicting both $i$ and $j$ with a lower confidence.

As the full distributional loss still outperforms classical cross entropy by 1.1 \%pt. on the full dataset, its other properties (such as the explicit margin and the independent logit evaluation) are likely to be beneficial for any dataset size.

\begin{table}[t]
\caption{Average test accuracies on the Cifar10 dataset when using different variations of the distributional loss.}
\label{tab:acc_ablation}
\centering
\begin{tabular}{l|llll}\hline
& \multicolumn{4}{l}{Samples per Class} \\
& 2                         & 20    & 200  & full dataset \\ \hline
full distributional  & \Chart{27.94}{20}{30}     & \Chart{55.23}{46}{56} & \Chart{71.20}{65}{75} & \Chart{84.41}{80}{90} \\
non-parametric   & \Chart{25.93}{20}{30}     & \Chart{52.97}{46}{56} & \Chart{68.67}{65}{75} & \Chart{84.13}{80}{90} \\
parametric   & \Chart{25.91}{20}{30}     & \Chart{55.10}{46}{56} & \Chart{68.65}{65}{75} & \Chart{84.24}{80}{90} \\
zero variance   & \Chart{23.65}{20}{30}     & \Chart{51.35}{46}{56} & \Chart{68.29}{65}{75} & \Chart{84.15}{80}{90} \\
CE baseline  & \Chart{23.50}{20}{30}     & \Chart{47.35}{46}{56} & \Chart{67.13}{65}{75} & \Chart{83.28}{80}{90} \\ \hline
\end{tabular}
\end{table}

\section{Conclusion}
This paper explores the idea of a distributional loss function: a classification loss that makes the distribution over all classifier outputs adhere to a specified structure rather than evaluating each sample and its corresponding label separately. The robustness of this distributional loss is most likely linked to a different bias imposed on the logit output of the network: it is assumed that the distribution of classification labels follows a Gaussian distribution, including samples with varying degrees of representativeness for their class, ranging from typical to atypical examples. This assumption is more reflective of natural classification tasks, and can be shown to reduce overfitting more effectively than cross entropy. 

While the proposed loss function comes at a cost of a moderate increase in training time, it constitutes an effective alternative for its intended purpose. In general, it improves classification accuracy, demonstrating superior performance over other recent loss formulations in the majority of evaluated scenarios. In settings with a high risk of overfitting, the test accuracy improved by up to 17 percentage points.

\begin{credits}

\subsubsection{\discintname}
The authors have no competing interests to declare that are relevant to the content of this article.
\end{credits}
%
%
%
\bibliographystyle{splncs04}
\bibliography{bibliography}
\end{document}

%% file: Figures/target_dist.tex
\definecolor{greyblue}{rgb}{0.54, 0.81, 0.94}
\definecolor{greygreen}{rgb}{0.55, 0.71, 0.}

\begin{subfigure}[t]{0.5\textwidth}
\begin{tikzpicture}
    \begin{axis}[
        axis lines = left, 
        height=3cm, width=6cm,
        axis y line=middle, 
        xtick={-1, -.5, 0, .5, 1},
        xticklabels={$-\infty$, $-m$, $0$, $m$, $-\infty$},
        ytick={},
        yticklabels={},
        x axis line style={white},
        ]
        \addplot[domain=-1:1, thick, color=greygreen, samples=300]{1/(0.05 * sqrt(2 * pi)) * exp(-0.5 * ((x + .5)^2)/(0.05^2))};
        \addplot[domain=-1:1, thick, color=greygreen, samples=300]{(1/(0.05 * sqrt(2 * pi)) * exp(-0.5 * ((x - .5)^2)/(0.05^2)))/4};

        \draw[color=red, thick] (axis cs:1, 0) -- (axis cs:1, 2);
        \draw[color=red, thick] (axis cs:-1, 0) -- (axis cs:-1, 15);

        \draw[dotted, thick] (axis cs:0.8, 0) -- (axis cs:1, 0);
        \draw[dotted, thick] (axis cs:-1, 0) -- (axis cs:-0.8, 0);
        \draw[thick] (axis cs:-0.8, 0) -- (axis cs:0.8, 0);
    \end{axis}
\end{tikzpicture}
\label{tikz:target_distB}
\caption{Logits $y_k(x)$ before softmax}
\end{subfigure}%
\begin{subfigure}[t]{0.5\textwidth}
\begin{tikzpicture}
    \begin{axis}[
        axis lines = left, 
        height=3cm, width=6cm,
        ytick={},
        yticklabels={},
        ]
        \addplot[domain=.05:1, thick, color=greygreen, samples=100]{exp(-(x-.05)^2/.01)};
        \addplot[domain=0:.95, thick, color=greygreen, samples=100]{exp(-(x-.95)^2/.01)/4};
        
        \addplot[domain=0:.05, thick, color=greygreen, samples=100]{((1 - (360 * (x-.05)^2))};
        \addplot[domain=.95:1, thick, color=greygreen, samples=100]{((1 - (360 * (x-.95)^2))/4};
        
        \draw[color=red, thick] (axis cs:1, 0) -- (axis cs:1, .25);
        \draw[color=red, thick] (axis cs:0.005, 0) -- (axis cs:0.005, 1);
    \end{axis}
\end{tikzpicture}
\label{tikz:target_distA}
\caption{Posterior probabilities $p_k(x)$ after softmax}
\end{subfigure}%